\documentclass{article} 
\usepackage{iclr2023_conference,times}

\usepackage{amsmath,amsfonts,bm}

\def\eqref#1{equation~\ref{#1}}

\def\1{\bm{1}}

\DeclareMathAlphabet{\mathsfit}{\encodingdefault}{\sfdefault}{m}{sl}
\SetMathAlphabet{\mathsfit}{bold}{\encodingdefault}{\sfdefault}{bx}{n}

\DeclareMathOperator*{\argmax}{arg\,max}

\usepackage{hyperref}
\usepackage{url}
\usepackage{graphicx}
\usepackage{amssymb}
\usepackage{wrapfig}

\title{Equivariant MuZero}

\author{Andreea Deac\thanks{Work performed while the author was at DeepMind.} \\
Mila -- Qu\'{e}bec Artificial Intelligence Institute\\
Universit\'{e} de Montr\'{e}al\\
\texttt{deacandr@mila.quebec} \\
\And
Th\'{e}ophane Weber \\
DeepMind \\
\texttt{theophane@deepmind.com} \\
\And
George Papamakarios \\
DeepMind\\
\texttt{gpapamak@deepmind.com}
}

\iclrfinalcopy 
\begin{document}

\maketitle

\begin{abstract}
	Deep reinforcement learning repeatedly succeeds in closed, well-defined domains such as games (Chess, Go, StarCraft). The next frontier is real-world scenarios, where setups are numerous and varied. For this, agents need to learn the underlying rules governing the environment, so as to robustly generalise to conditions that differ from those they were trained on. Model-based reinforcement learning algorithms, such as the highly successful MuZero, aim to accomplish this by learning a world model. However, leveraging a world model has not consistently shown greater generalisation capabilities compared to model-free alternatives. In this work, we propose improving the data efficiency and generalisation capabilities of MuZero by explicitly incorporating the \emph{symmetries} of the environment in its world-model architecture. We prove that, so long as the neural networks used by MuZero are equivariant to a particular symmetry group acting on the environment, the entirety of MuZero's action-selection algorithm will also be equivariant to that group. We evaluate Equivariant MuZero on procedurally-generated MiniPacman and on Chaser from the ProcGen suite: training on a set of mazes, and then testing on unseen rotated versions, demonstrating the benefits of equivariance. Further, we verify that our performance improvements hold even when only some of the components of Equivariant MuZero obey strict equivariance, which highlights the robustness of our construction.
\end{abstract}

\section{Introduction}
Reinforcement learning (RL) is a potent paradigm for solving sequential decision making problems in a dynamically changing environment. Successful examples of its uses include game playing \citep{vinyals2019grandmaster}, drug design \citep{segler2018planning}, robotics \citep{ibarz2021train} and theoretical computer science \citep{fawzi2022discovering}. However, the generality of RL often leads to data inefficiency, poor generalisation to situations that differ from those encountered in training, and lack of safety guarantees. This is an issue especially in domains where data is scarce or difficult to obtain, such as medicine or human-in-the-loop scenarios. 

Most RL approaches do not directly attempt to capture the regularities present in the environment. As an example, consider a grid-world: moving down in a maze is equivalent to moving left in the $90^{\circ}$ clock-wise rotation of the same maze.
Such equivalences can be formalised via Markov Decision Process homomorphisms \citep{ravindran2004algebraic, ravindran2004approximate}, and while some works incorporate them \citep[e.g.][]{van2020mdp, rezaei2022continuous}, 
most deep reinforcement learning agents would act differently in such equivalent states if they do not observe enough data. This becomes even more problematic when the number of equivalent states is large. One common example is 3D regularities, such as changing camera angles in robotic tasks.

In recent years, there has been significant progress in building deep neural networks that explicitly obey such regularities, often termed geometric deep learning \citep{bronstein2021geometric}. In this context, the regularities are formalised using symmetry groups and architectures are built by composing transformations that are equivariant to these symmetry groups (e.g.\ convolutional neural networks for the translation group, graph neural networks and transformers for the permutation group).

As we are looking to capture the symmetries present in an environment, a fitting place is within the framework of model-based RL (MBRL). MBRL leverages explicit world-models to forecast the effect of action sequences, either in the form of next-state or immediate reward predictions. These imagined trajectories are used to construct plans that optimise the forecasted returns. In the context of state-of-the-art MBRL agent MuZero \citep{schrittwieser2020mastering}, a Monte-Carlo tree search is executed over these world-models in order to perform action selection. 

\begin{figure}
    \centering
    \includegraphics[width=.9\linewidth]{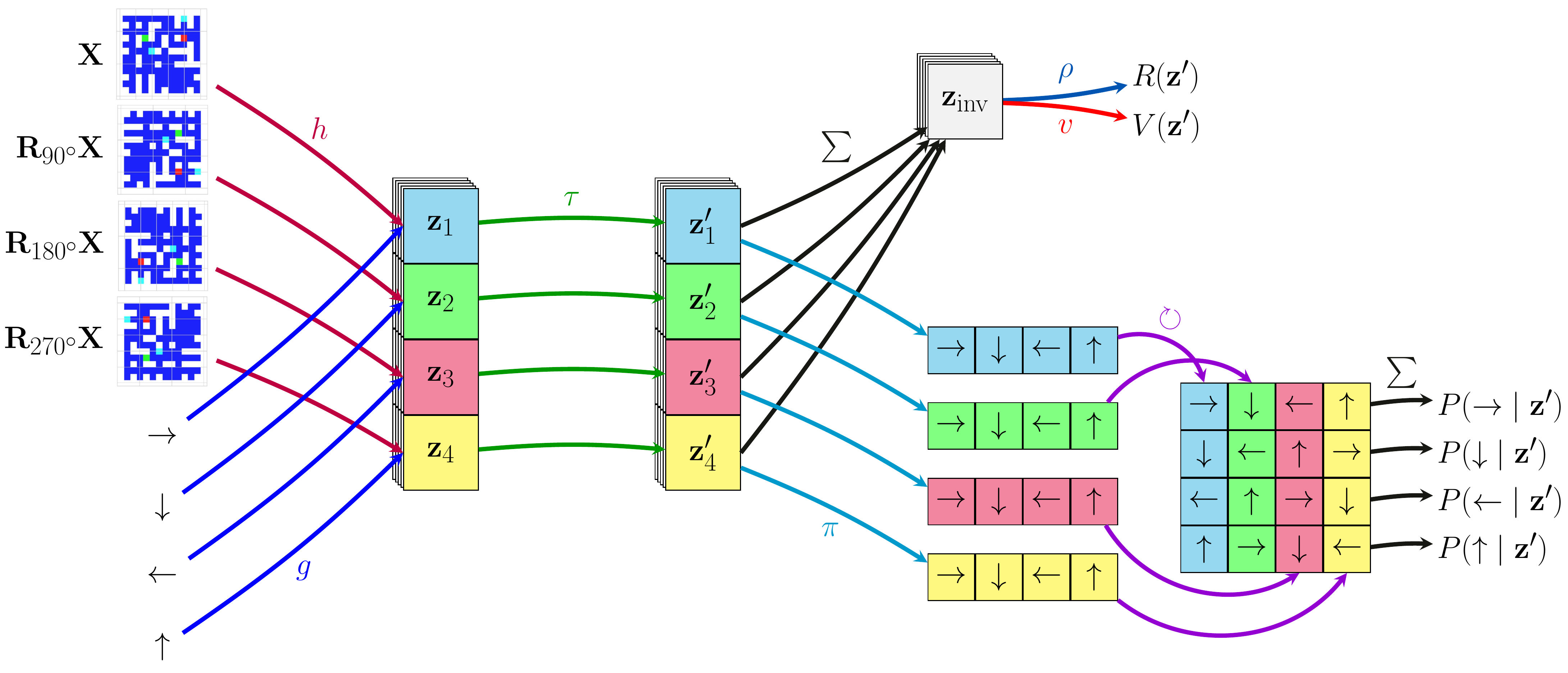}
    \caption{Architecture of Equivariant MuZero, where $h$, $g$ are encoders, $\tau$ is the transition model, $\rho$ is the reward model, $v$ is the value model and $\pi$ is the policy predictor. Each colour represents an element of the $C_4$ group $\{{\bf I}, {\bf R}_{90^\circ}, {\bf R}_{180^\circ}, {\bf R}_{270^\circ}\}$ applied to the input (observation and action).}
    \label{fig:eqmz}
\end{figure}

In this paper, we demonstrate that equivariance and MBRL can be effectively combined by proposing Equivariant MuZero (EqMuZero, shown in Figure \ref{fig:eqmz}), a variant of MuZero where equivariance constraints are enforced by design in its constituent neural networks. As MuZero does not use these networks directly to act, but rather executes a search algorithm on top of their predictions, it is not immediately obvious that the actions taken by the EqMuZero agent would obey the same constraints---is it guaranteed to produce a rotated action when given a rotated maze? One of our key contributions is a proof that guarantees this: as long as all neural networks are equivariant to a symmetry group, all actions taken will also be equivariant to that same symmetry group. Consequently, EqMuZero can be more data-efficient than standard MuZero, as it knows by construction how to act in states it has never seen before. We empirically verify the generalisation capabilities of EqMuZero in two grid-worlds: procedurally-generated MiniPacman and the Chaser game in the ProcGen suite.

\section{Background}
\paragraph{Reinforcement Learning}
The reinforcement learning problem is typically formalised as a Markov Decision Process $(S, A, P, R, \gamma)$ formed from a set of states $S$, a set of actions $A$, a discount factor $\gamma\in[0, 1]$, and two functions that model the outcome of taking action $a$ in state $s$: the transition distribution $P(s'|s, a)$---specifying the next state probabilities---and the reward function $R(s, a)$---specifying the expected reward. The aim is to learn a \emph{policy}, $\pi(a|s)$, a function specifying (probabilities of) actions to take in state $s$, such that the agent maximises the (expected) cumulative reward $G(\tau)=\sum_{t=0}^{t=T}\gamma^t R(s_t, a_t)$, where $\tau=(s_0, a_0, s_1, a_1, \ldots, s_T, a_T)$ is the trajectory taken by the agent starting in the initial state $s_0$ and following the policy to decide $a_t$ based on $s_t$.

\paragraph{MuZero} 
Reinforcement learning agents broadly fall into two categories: \emph{model-free} and \emph{model-based}. The specific agent we extend here, MuZero \citep{schrittwieser2020mastering}, is a model-based agent for deterministic environments (where $P(s' | s, a) = 1$ for exactly one $s'$ for all $s\in S$ and $a\in A$). MuZero relies on several neural-network components that are composed to create a \emph{world model}. These components are: the \emph{encoder}, $E : S \rightarrow Z$, which embeds states into a latent space $Z$ (e.g.\ $Z = \mathbb{R}^k$), the \emph{transition model}, $T : Z \times A \rightarrow Z$, which predicts embeddings of next states, the \emph{reward model}, $R : Z\times A\rightarrow \mathbb{R}$, which predicts the immediate expected reward after taking an action in a particular state, the \emph{value model}, $V : Z\rightarrow \mathbb{R}$, which predicts the value (expected cumulative reward) from this state, and the \emph{policy model} $P : Z\rightarrow [0, 1]^{|A|}$, which predicts the probability of taking each action from the current state. To plan its next action, MuZero executes a Monte Carlo tree search (MCTS) over many simulated trajectories, generated using the above models.

MuZero has demonstrated state-of-the-art capabilities over a variety of deterministic or near-deterministic environments, such as Go, Chess, Shogi and Atari, and has been successfully applied to real-world domains such as video compression \citep{mandhane2022muzero}. Although here we focus on MuZero for deterministic environments, we note that extensions to stochastic environments also exist \citep{antonoglou2021planning} and are an interesting target for future work.

\paragraph{Groups and Representations} A \emph{group} $(\mathfrak{G}, \circ)$ is a set $\mathfrak{G}$ equipped with a \emph{composition} operation $\circ : \mathfrak{G} \times \mathfrak{G} \rightarrow \mathfrak{G}$ (written concisely as $\mathfrak{g}\circ \mathfrak{h} = \mathfrak{g}\mathfrak{h}$), satisfying the following axioms: \emph{(associativity)} $(\mathfrak{g}\mathfrak{h})\mathfrak{l} = \mathfrak{g}(\mathfrak{h}\mathfrak{l})$ for all $\mathfrak{g}, \mathfrak{h}, \mathfrak{l}\in\mathfrak{G}$;  \emph{(identity)} there exists a unique $\mathfrak{e}\in\mathfrak{G}$ satisfying $\mathfrak{e}\mathfrak{g}=\mathfrak{g}\mathfrak{e}=\mathfrak{g}$ for all $\mathfrak{g}\in\mathfrak{G}$; \emph{(inverse)} for every $\mathfrak{g}\in\mathfrak{G}$ there exists a unique $\mathfrak{g}^{-1}\in\mathfrak{G}$ such that $\mathfrak{g}\mathfrak{g}^{-1} = \mathfrak{g}^{-1}\mathfrak{g}=\mathfrak{e}$.

Groups are a natural way to describe \emph{symmetries}: object transformations that leave them unchanged. They can be reasoned about in the context of linear algebra by using their \emph{real representations}: functions $\rho_\mathcal{V} : \mathfrak{G}\rightarrow\mathbb{R}^{N\times N}$ that give, for every group element $\mathfrak{g}\in\mathfrak{G}$, a real matrix demonstrating how this element \emph{acts} on a vector space $\mathcal{V}$. For example, for the rotation group $\mathfrak{G}=\mathrm{SO}(n)$, the representation $\rho_\mathcal{V}$ would provide an appropriate $n\times n$ rotation matrix for each rotation $\mathfrak{g}$.

\paragraph{Equivariance and Invariance} As symmetries are assumed to not change the essence of the data they act on, we would like to construct neural networks that adequately represent such symmetry-transformed inputs. Assume we have a neural network $f : \mathcal{X} \rightarrow \mathcal{Y}$, mapping between vector spaces $\mathcal{X}$ and $\mathcal{Y}$, and that we would like this network to respect the symmetries within a group $\mathfrak{G}$. Then we can impose the following condition, for all group elements $\mathfrak{g}\in\mathfrak{G}$ and inputs $\mathbf{x}\in\mathcal{X}$:
\begin{equation}
    f(\rho_\mathcal{X}(\mathfrak{g})\mathbf{x}) = \rho_\mathcal{Y}(\mathfrak{g})f(\mathbf{x}).
\end{equation}
This condition is known as \emph{$\mathfrak{G}$-equivariance}---for any group element, it does not matter whether we act with it on the input or on the output of the function $f$---the end result is the same. A special case of this, \emph{$\mathfrak{G}$-invariance}, is when the output representation is trivial ($\rho_\mathcal{Y}(\mathfrak{g})={\bf I}$): 
\begin{equation}
  f(\rho_\mathcal{X}(\mathfrak{g})\mathbf{x}) = f(\mathbf{x}).
\end{equation}
In geometric deep learning, equivariance to reflections, rotations, translations and permutations has been of particular interest \citep{bronstein2021geometric}. 

Generally speaking, there are three ways to obtain an equivariant model: a) data augmentation, b) data canonicalisation and c) specialised architectures. 
Data augmentation creates additional training data by applying group elements $\mathfrak{g}$ to input/output pairs $(\mathbf{x}, \mathbf{y})$---equivariance is encouraged by training on the transformed data and/or minimising auxiliary losses such as $\|\rho_\mathcal{Y}(\mathfrak{g})f(\mathbf{x}) - f(\rho_\mathcal{X}(\mathfrak{g})\mathbf{x})\|$. Data augmentation can be simple to apply, but it results in only approximate equivariance.
Data canonicalisation requires a method to standardise the input, such as breaking the translation symmetry for molecular representation by centering the atoms around the origin \citep{musil2021physics}---however, in many cases, such as the relatively simple MiniPacman environment we use in our experiments, such a canonical transformation may not exist.
Specialised architectures have the downside of being harder to build, but they can guarantee exact equivariance---as such, they reduce the search space of functions, potentially reducing the number of parameters and increasing training efficiency.

\begin{wrapfigure}[23]{R}[0pt]{0.39\textwidth}
\begin{minipage}{0.39\textwidth}
    \centering
	\includegraphics[scale=.22]{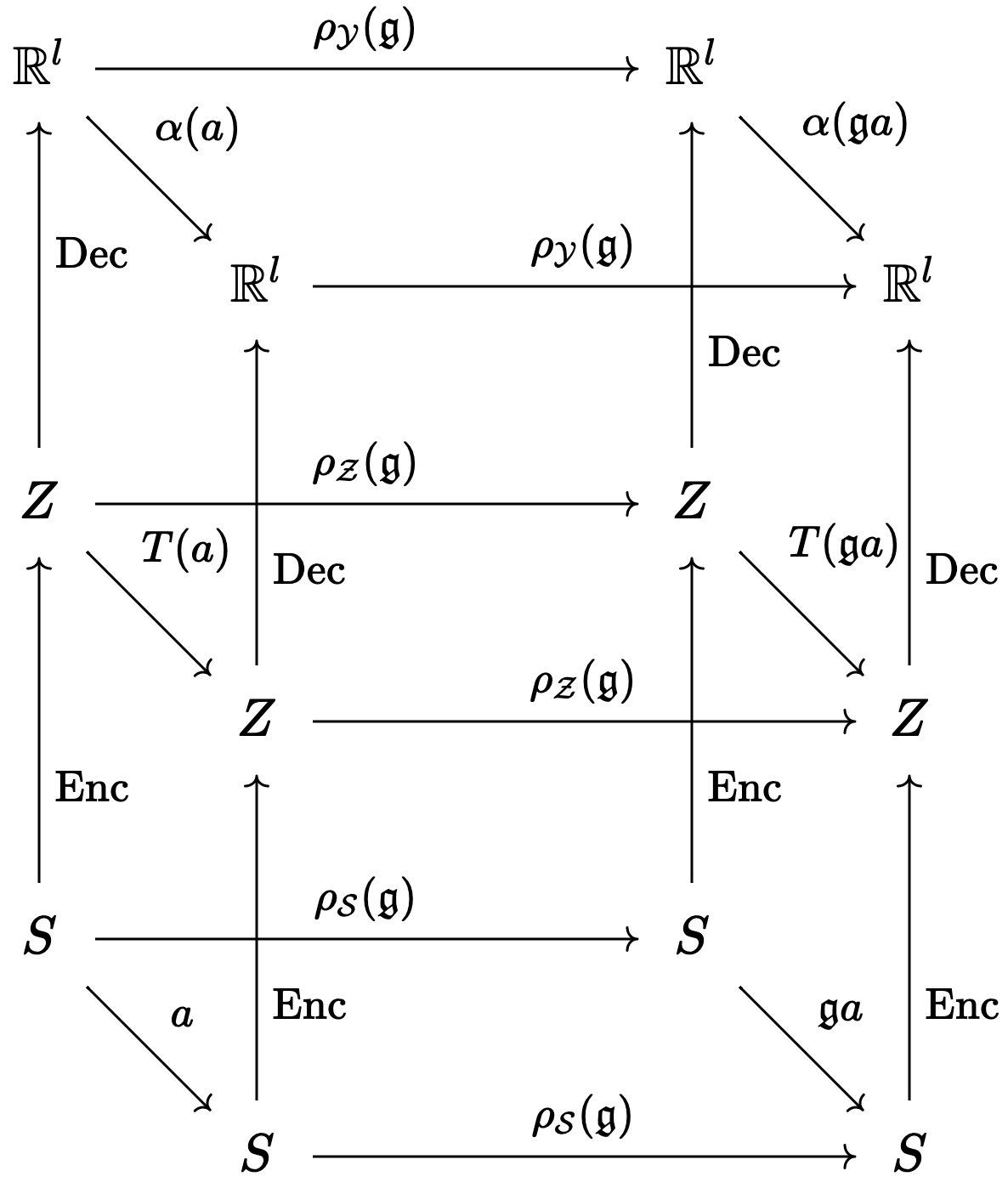}
	\caption{Commutative diagram of symmetries in RL\@. State transitions due to an action $a$ are back-to-front, transformations due to a symmetry $\mathfrak{g}$ are left-to-right, state encoding and decoding by the model is bottom-to-top.}
	\label{fig:eq_cube}
\end{minipage}
\end{wrapfigure}

\paragraph{Equivariance in RL} 
There has been previous work at the intersection of reinforcement learning and equivariance. While leveraging multi-agent symmetries was repeatedly shown to hold promise \citep{van2021multi, muglich2022equivariant}, of particular interest to us are the symmetries emerging from the environment, in a single-agent scenario. Related work in this space can be summarised by the commutative diagram in Figure \ref{fig:eq_cube}.
When considering only the cube at the bottom, we recover \cite{park2022learning}---a supervised learning task where a latent transition model $T$ learns to predict the next state embedding. They show that if $T$ is equivariant, the encoder can pick up the symmetries of the environment even if it is not fully equivariant by design. \citet{mondal2022eqr} build a model-free agent by combining an equivariant-by-design encoder and enforcing the remaining equivariances via regularisation losses. They also consider the invariance of the reward, captured in Figure \ref{fig:eq_cube} by taking the decoder to be the reward model and $l=1$. The work of \cite{van2020mdp} can be described by having the value model as the decoder, while the work of \cite{wang2022mathrm} has the decoder as the policy model and $l=|A|$.

\section{Experiments and results}

\paragraph{Environments}
We consider two 2D grid-world
environments, MiniPacman \citep{guez2019investigation} and Chaser \citep{cobbe2020leveraging}, that feature an agent navigating in a 2D maze. In both environments, the state is the grid-world map $\mathbf{X}$ and an action is a direction to move. Both of these grid-worlds are symmetric with respect to $90 ^{\circ}$ rotations, in the sense that moving down in some map is the same as moving left in the $90^\circ$ clock-wise rotated version of the same map. Hence, we take our symmetry group to be $\mathfrak{G}=C_4=\{{\bf I}, {\bf R}_{90^\circ}, {\bf R}_{180^\circ}, {\bf R}_{270^\circ}\}$, the 4-element cyclic group, which in our case represents rotating the map by all four possible multiples of $90^{\circ}$.

\paragraph{Equivariant MuZero}
In what follows, we describe how the various components of EqMuZero (Figure \ref{fig:eqmz}) are designed to obey $C_4$-equivariance. For simplicity, we assume there are only four directional movement actions in the environment ($A = \{\rightarrow, \downarrow, \leftarrow, \uparrow\}$). Any additional non-movement actions (such as the ``do nothing'' action) can be included without difficulty.

To enforce $C_4$-equivariance in the encoder, we first need to specify the effect of rotations on the latent state $\mathbf{z}$. In our implementation, the latent state consists of 4 equally shaped arrays, $\mathbf{z} = ({\bf z}_1, {\bf z}_2, {\bf z}_3,{\bf z}_4)$, and we prescribe that a $90^\circ$ clock-wise rotation manifests as a cyclical permutation: $\mathbf{R}_{90^\circ}\mathbf{z} = ({\bf z}_2, {\bf z}_3, {\bf z}_4,{\bf z}_1)$. Then, our equivariant encoder embeds state $\bf X$ and action $a$ as follows:
\begin{equation}\label{eqenc}
    E({\bf X}, a) = (h({\bf X}) + g(a), h({\bf R}_{90^\circ}\!{\bf X}) + g({\bf R}_{90^\circ}a), h({\bf R}_{180^\circ}\!{\bf X}) + g({\bf R}_{180^\circ}a), h({\bf R}_{270^\circ}\!{\bf X}) + g({\bf R}_{270^\circ}a))
\end{equation}
where $h$ is a CNN and $g$ is an MLP\@. For the summation, the output of $g$ is accordingly broadcasted across all pixels of $h$'s output.
It is easy to verify that this equation satisfies $C_4$-equivariance, that is, $E({\bf R}_{90^\circ}{\bf X}, {\bf R}_{90^\circ}a) = {\bf R}_{90^\circ}E({\bf X}, a)$.

We can build a $C_4$-equivariant transition model by maintaining the structure in the latent space:
\begin{equation}\label{eq:eqTS}
    T({\bf z}) = (\tau({\bf z}_1), \tau({\bf z}_2), \tau({\bf z}_3), \tau({\bf z}_4)). 
\end{equation}
It is also possible to have a less constrained transition model that allows components of ${\bf z}$ to \emph{interact}, while still retaining $C_4$-equivariance, as follows:
\begin{equation}\label{eq:eqT}
    T({\bf z}) = (\tau({\bf z}_1, {\bf z}_2, {\bf z}_3,{\bf z}_4), \tau({\bf z}_2, {\bf z}_3, {\bf z}_4,{\bf z}_1), \tau({\bf z}_3, {\bf z}_4, {\bf z}_1,{\bf z}_2), \tau({\bf z}_4, {\bf z}_1, {\bf z}_2,{\bf z}_3)).
\end{equation}
In our experiments, we use the more constrained variant for MiniPacman, and the less constrained variant for Chaser, as more data is available for the latter. In either case, we take $\tau$ to be a ResNet.

The policy model is made $C_4$-equivariant by appropriately combining state and action embeddings from all four latents in $\bf{z}$:
\begin{equation}\label{eq:eqP}
    P(a\,|\,{\bf z}) = \frac{\pi(a\,|\,{\bf z}_1) + \pi({\bf R}_{90^\circ}a\,|\,{\bf z}_2) + \pi({\bf R}_{180^\circ}a\,|\,{\bf z}_3) + \pi({\bf R}_{270^\circ}a\,|\,{\bf z}_4)}{4}
\end{equation}
where $\pi(\cdot\,|\,{\bf z}_i)$ is an MLP followed by a softmax, which produces a probability distribution over actions given the map encoded by ${\bf z}_i$. It is easy to show that $\sum_{a\in A} P(a\,|\,{\bf z}) = 1$, i.e.\ $P(\cdot\,|\,{\bf z})$ is properly normalised, and that $P({\bf R}_{90^\circ}a\,|\,{\bf R}_{90^\circ}{\bf z}) = P(a\,|\,{\bf z})$, i.e.\ it satisfies $C_4$-equivariance.

Lastly, the reward and value networks ($R$, $V$), modeled by MLPs $\rho$ and $v$ respectively,  should be $C_4$-invariant. We can satisfy this constraint by \emph{aggregating} the latent space with any $C_4$-invariant function, such as sum, average or max. Here we use summation:
\begin{equation}\label{eq:invRV}
    R({\bf z}) = \rho({\bf z}_1 + {\bf z}_2 + {\bf z}_3 + {\bf z}_4), \qquad V({\bf z}) = v({\bf z}_1 + {\bf z}_2 + {\bf z}_3 + {\bf z}_4).
\end{equation}

Composing the equivariant components described above (Equations \ref{eqenc}--\ref{eq:invRV}), we construct the end-to-end equivariant EqMuZero agent, displayed in Figure \ref{fig:eqmz}. In appendix \ref{app:proof}, we prove that, assuming that all the relevant neural networks used by MuZero are $\mathfrak{G}$-equivariant, the proposed EqMuZero agent will select actions in a $\mathfrak{G}$-equivariant manner.

\paragraph{Results}
We compare EqMuZero with a standard MuZero that uses non-equivariant components: ResNet-style networks for the encoder and transition models, and MLP-based policy, value and reward models, following \cite{hamrick2020role}. Moreover, as the encoder and the policy of EqMuZero are the only two components which require knowledge of how the symmetry group acts on the environment, we include the following ablations in order to evaluate the trade-off between end-to-end equivariance and general applicability: Standard MuZero with an equivariant encoder, equivariant MuZero with a standard encoder and equivariant MuZero with a standard policy model.

\begin{figure}
	\centering
	\hspace{-6pt}\includegraphics[width=\linewidth]{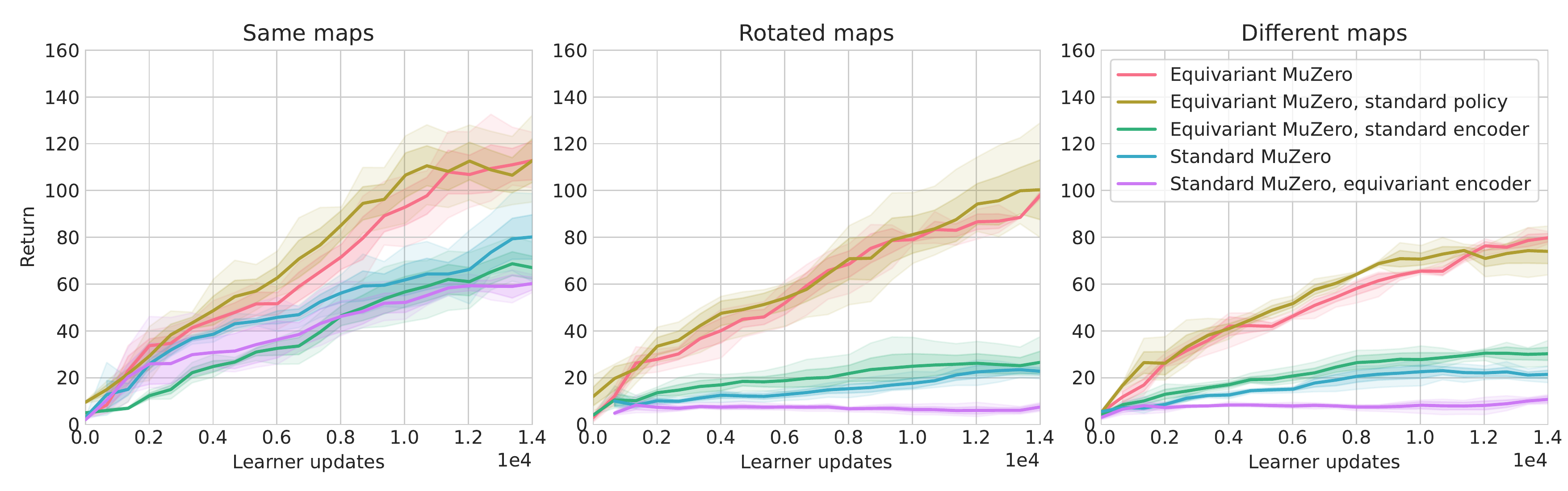}\\
	\includegraphics[width=\linewidth]{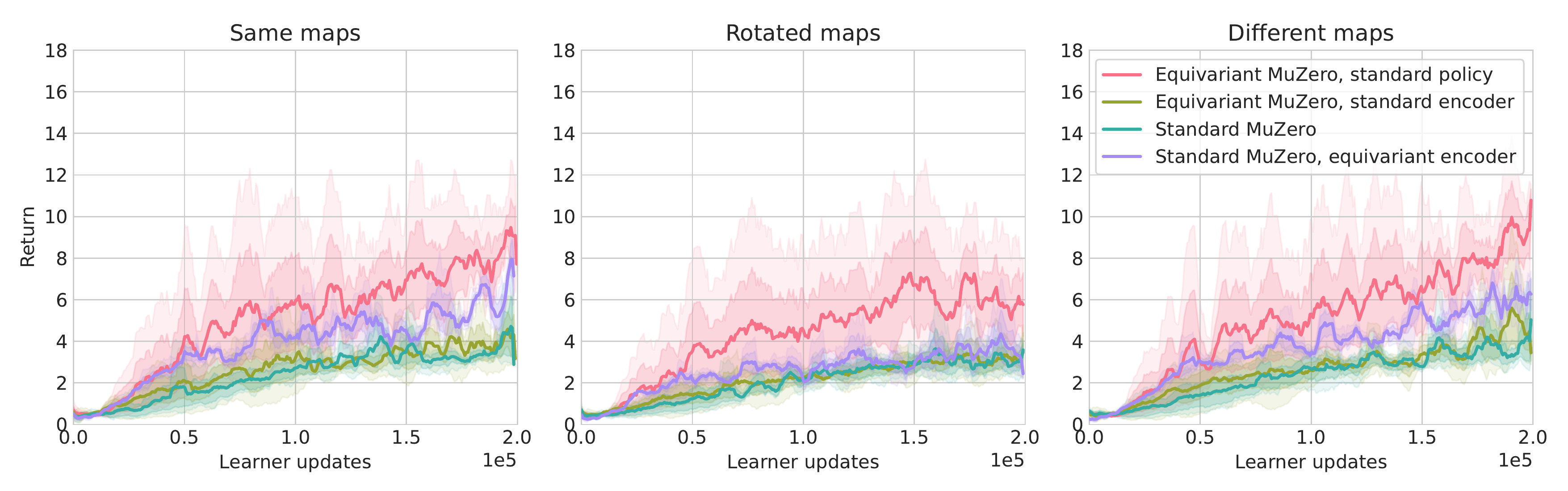}
	\caption{Results on procedurally-generated MiniPacman (top) and Chaser from ProcGen (bottom).}
	\label{fig:results}
\end{figure}

We train each agent on a set of maps, ${\bf X}$. 
To test for generalisation, we measure the agent's performance on three, progressively harder, settings. Namely, we evaluate the agent on ${\bf X}$, with randomised initial agent position (denoted by \emph{same} in our results), on the set of rotated maps ${\bf RX}$, where ${\bf R}\in\{{\bf R}_{90^\circ}, {\bf R}_{180^\circ}, {\bf R}_{270^\circ}\}$ (denoted by \emph{rotated}) and, lastly, on a set of maps $\bf Y$, such that $ {\bf Y}  \cap {\bf X} =  \varnothing $ and $ {\bf Y}  \cap {\bf RX} = \varnothing$ (denoted by \emph{different}).

Figure \ref{fig:results} (top) presents the results of the agents on MiniPacman. First, we empirically confirm that the average reward on layouts $\bf{X}$, seen during training, matches the average reward gathered on the rotations of the same mazes, $\bf{RX}$, for EqMuZero. 
Second, we notice that changing the equivariant policy with a non-equivariant one does not significantly impact performance. 
However, the same swap in the encoder brings the performance of the agent down to that of Standard MuZero---this suggests that the structure in the latent space of the transition model, when not combined with some explicit method of imposing equivariance in the encoder, does not provide noticeable benefits.
Third, we notice that Equivariant MuZero is generally robust to layout variations, as the learnt high-reward behaviours also transfer to $\bf{Y}$. At the same time, Standard MuZero significantly drops in performance for both $\bf{Y}$ and $\bf{RX}$. We note that experiments on MiniPacman were done in a low-data scenario, using 5 maps of size $14 \times 14$ for training; we observed that the differences between agents diminished when all agents were trained with at least 20 times more maps.

Figure \ref{fig:results} (bottom) compares the performance of the agents on the ProcGen game, Chaser, which has similar dynamics to MiniPacman, but larger mazes of size $ 64 \times 64$  and a more complex action space. Due to the complexity of the action space, we only use EqMuZero with a standard policy, rather than a fully equivariant version. We use 500 maze instances for training. Our results demonstrate that, even when the problem complexity is increased in such a way, Equivariant MuZero still consistently outperforms the other agents, leading to more robust plans being discovered.

\bibliography{iclr2023_conference}
\bibliographystyle{iclr2023_conference}

\appendix

\section{Proof of MuZero Equivariance} \label{app:proof}
Assume our neural networks are: $h$ for the encoder, $\tau$ for the transition model, $\pi$ for the policy model, $v$ for the value model and $\rho$ for the reward model. By design, we make $h, \tau  \text{ and } \pi$ be $\mathfrak{G}$-equivariant, and $v$ and $\rho$ be $\mathfrak{G}$-invariant.

The reward, value, policy and transition respect the equivariances, as compositions of equivariant functions:
\begin{equation}
\begin{aligned}
    R &= \rho \tau ^ k h \\
    V &= v \tau ^ k h \\
    P &= \pi \tau ^ k h \\
    T &= \tau ^k h.
\end{aligned}
\end{equation}
Then, the return is also a $\mathfrak{G}$-invariant function as it is the sum of two $\mathfrak{G}$-invariant functions:
\begin{equation}
G(s^k) = \sum_{\tau=0}^{l-1-k}{\gamma ^ \tau} \rho(s^{k+\tau}, a^{k+1+\tau}) + \gamma^{l - k} v(s^l, a^{l+1}).
\end{equation}

For proving that one planning step is equivariant, we need to show that the action selection is $\mathfrak{G}$-equivariant. 

Since the outcome of MuZero's MCTS function is based on the initial observation, $o$, we denote MCTS's internal state as $\{Q^o(s, a), N^o(s, a), \ldots\}$. We use identical notation as \citet{schrittwieser2020mastering} for these states, even though we express the MuZero models $R,V,P,T$ somewhat differently.

Knowing how they are updated:
\begin{equation} \label{eq:action}
a^{k} \!= \arg\!\max_{a}\!\bigg[
Q^o(s^{k-1}\!, a) + P^o(s^{k-1}\!, a) \frac{\sqrt{\sum_b N^o(s^{k-1}\!, b)}}{1 + N^o(s^{k-1}\!, a)} \bigg(\!c_1 + \log\!\Big(\frac{\sum_b \!N^o(s^{k-1}\!, b) + c_2 + 1}{c_2}\Big)\!\bigg)\! \bigg]
\end{equation}
\begin{equation} \label{eq:QN}
\begin{aligned}
Q^o_t(s^{k-1},a^k) & = \frac{N^o_{t-1}(s^{k-1},a^k) Q^o_{t-1}(s^{k-1},a^k) + G(s^{k-1})}{N^o_{t-1}(s^{k-1},a^k) + 1} \\
N^o_t(s^{k-1},a^k) & = N^o_{t-1}(s^{k-1},a^k) + 1.
\end{aligned}
\end{equation}

As discussed previously, we need to show that, for each MCTS internal state (e.g.\ $N^o$), if we assume $\pi, v, \tau, \rho, h$ to be equivariant functions, the resulting state would also be equivariant under transformations of the initial observation. That is, for all $s, a$:
\begin{equation}
N^{\mathfrak{g}_oo}(\mathfrak{g}_ss, \mathfrak{g}_aa) = N^o(s, a).
\end{equation}

To prove this, we will use induction on the number of backups performed by MCTS, $t$. We proceed:
\begin{equation}
\begin{aligned}
    \text{Base case } (t=0): N_0^{\mathfrak{g}_oo}(\mathfrak{g}_ss,\mathfrak{g}_aa) &= N_0^o(s, a)=0 \\
    Q_0^{\mathfrak{g}_oo}(\mathfrak{g}_ss,\mathfrak{g}_aa) &= Q_0^o(s, a)=0.
\end{aligned}
\end{equation}
Assume: 
\begin{equation}
\begin{aligned}
 \text{Case } t: N^{\mathfrak{g}_oo}_t(\mathfrak{g}_ss,\mathfrak{g}_aa) &= N^o_t(s, a) \\ Q^{\mathfrak{g}_oo}_t(\mathfrak{g}_ss,\mathfrak{g}_aa) &= Q^o_t(s, a).  
\end{aligned}
\end{equation}

We will start by showing that the states and actions expanded by MCTS under initial $\mathfrak{G}$-transformed observation $\mathfrak{g}_oo$ $(\widetilde{s}^0, \widetilde{a}^1, \widetilde{s}^1, \widetilde{a}^2, \dots)$, would exactly correspond to $(\mathfrak{g}_ss^0, \mathfrak{g}_aa^1, \mathfrak{g}_ss^1, \mathfrak{g}_aa^2, \dots)$, where $(s^0, a^1, s^1, a^2, \dots)$ are states expanded under the non-transformed observation, $o$.

By equivariance of $h$, $\widetilde{s}^0 = h(\mathfrak{g}_oo)=\mathfrak{g}_sh(o) = \mathfrak{g}_ss^0$, as expected.

Next, we show that the actions selected by MCTS also obey a $\mathfrak{G}$-equivariance constraint, in the sense that: if $\widetilde{s}^{k-1} = \mathfrak{g}_ss^{k-1}$, then $\widetilde{a}^k=\mathfrak{g}_aa^k$.

As we assumed $N_t^o$ to be $\mathfrak{G}$-equivariant, it must hold that $\sum_b N_t^o(s, b)$ is $\mathfrak{G}$-invariant (as a sum-reduction of equivariant functions). Hence, we can rewrite Equation \ref{eq:action} as: 
\begin{equation} \label{pUCT}
a^{k} = \arg\max_{a}\bigg[
Q^o_t(s^{k-1}, a) + P^o_t(s^{k-1}, a) \frac{\epsilon(s^{k-1})}{1 + N^o_t(s^{k-1}, a)} \bigg]
\end{equation}
where $\epsilon$ is $\mathfrak{G}$-invariant, $P^o$ is $\mathfrak{G}$-equivariant by composition of functions that are $\mathfrak{G}$-equivariant by assumption, and $Q^o$ is $\mathfrak{G}$-equivariant by assumption of $ \text{Case } t$.

Hence, using this formula to define $\widetilde{a}^k$, we recover:
\begin{align*}
    \widetilde{a}^k &= \arg\max_a\bigg[
Q^{\mathfrak{g}_oo}_t(\widetilde{s}^{k-1}, a) + P^{\mathfrak{g}_oo}_t(\widetilde{s}^{k-1}, a) \frac{\epsilon(\widetilde{s}^{k-1})}{1 + N^{\mathfrak{g}_oo}_t(\widetilde{s}^{k-1}, a)} \bigg]\\
 &= \arg\max_a\bigg[
Q^{\mathfrak{g}_oo}_t(\mathfrak{g}_ss^{k-1}, a) + P^{\mathfrak{g}_oo}_t(\mathfrak{g}_ss^{k-1}, a) \frac{\epsilon(\mathfrak{g}_ss^{k-1})}{1 + N^{\mathfrak{g}_oo}_t(\mathfrak{g}_ss^{k-1}, a)} \bigg]\\
 &= \arg\max_a\bigg[
Q^{\mathfrak{g}_oo}_t(\mathfrak{g}_ss^{k-1}, \mathfrak{g}_a\mathfrak{g}^{-1}_aa) + P^{\mathfrak{g}_oo}_t(\mathfrak{g}_ss^{k-1}, \mathfrak{g}_a\mathfrak{g}^{-1}_aa) \frac{\epsilon(\mathfrak{g}_ss^{k-1})}{1 + N^{\mathfrak{g}_oo}_t(\mathfrak{g}_ss^{k-1}, \mathfrak{g}_a\mathfrak{g}^{-1}_aa)} \bigg]\\
 &= \arg\max_a\bigg[
Q^{o}_t(s^{k-1},\mathfrak{g}^{-1}_aa) + P^{o}_t(s^{k-1},\mathfrak{g}^{-1}_aa) \frac{\epsilon(s^{k-1})}{1 + N^{o}_t(s^{k-1},\mathfrak{g}^{-1}_aa)} \bigg]\\
 &= \mathfrak{g}_a\arg\max_a\bigg[
Q^{o}_t(s^{k-1},a) + P^{o}_t(s^{k-1},a) \frac{\epsilon(s^{k-1})}{1 + N^{o}_t(s^{k-1},a)} \bigg] \\
&= \mathfrak{g}_aa^k.
\end{align*}
Note that we have taken the $\mathfrak{g}_a$ out of the $\argmax$, which is an unambiguous operation only if there is a unique action $a^k$ that maximises the expression in Equation \ref{pUCT}. To avoid breaking the symmetry in practice, we propose that tiebreaks for $a^k$ are resolved in a purely randomised fashion.

Showing this, we now only need to verify that the updates to $N_t$ and $Q_t$ (in Equation \ref{eq:QN})
are equivariant for all state-action pairs along the trajectory. Values of $N$ and $Q$ for all other state-action pairs will be unchanged from $N_t$, and therefore trivially still $\mathfrak{G}$-equivariant.

First we show this for $N$:
\begin{align*}
    N_{t+1}^{\mathfrak{g}_oo}(\widetilde{s}^{k-1}, \widetilde{a}^{k}) &= N_{t+1}^{\mathfrak{g}_oo}(\mathfrak{g}_ss^{k-1},\mathfrak{g}_aa^k) \\&= N_{t}^{\mathfrak{g}_oo}(\mathfrak{g}_ss^{k-1}, \mathfrak{g}_aa^k) + 1 &\\ &= N^o_t(s^{k-1}, a^k) + 1\\ &= N^o_{t+1}(s^{k-1}, a^k).
\end{align*}
Hence, $ \text{Case } t+1$ still holds for $N$. Now we turn our attention to $Q$.

First, by invariance of $\rho$ and $w$, we can show that $G(s^k)$ is a sum of $\mathfrak{G}$-invariant functions and therefore also invariant. Plugging into the $Q$ update:
\begin{align*}
    Q_{t+1}^{\mathfrak{g}_oo}(\widetilde{s}^{k-1}, \widetilde{a}^k)&= Q_{t+1}^{\mathfrak{g}_oo}(\mathfrak{g}_ss^{k-1},\mathfrak{g}_aa^k)\\
    &=\frac{N^{\mathfrak{g}_oo}_{t}(\mathfrak{g}_ss^{k-1},\mathfrak{g}_aa^k) Q^{\mathfrak{g}_oo}_{t}(\mathfrak{g}_ss^{k-1},\mathfrak{g}_aa^k) + G(\mathfrak{g}_ss^{k-1})}{N^{\mathfrak{g}_oo}_{t}(\mathfrak{g}_ss^{k-1},\mathfrak{g}_aa^k) + 1}\\
    &=\frac{N^o_{t}(s^{k-1},a^k) Q^{o}_{t}(s^{k-1},a^k) + G(s^{k-1})}{N^{o}_{t}(s^{k-1},a^k) + 1} \\&= Q_{t+1}^o(s^{k-1}, a^k).
\end{align*}
Hence, $ \text{Case } t+1$ also holds for $Q$. As discussed before, we assume it holds by composition for all other state stored by MCTS ($P, T, R$).

Having proved that all internal state of of MCTS consistently remains transformed by $\mathfrak{G}$ under transformed input observations, we can conclude that the final policy given by MCTS will be exactly $\mathfrak{G}$-equivariant.

\end{document}